\documentclass[runningheads]{llncs}
\usepackage[T1]{fontenc}
\usepackage{graphicx}
\usepackage{hyperref}
\usepackage{color}

\usepackage{amsmath}
\usepackage{amssymb}

\usepackage{cleveref}
\usepackage{todonotes}
\usepackage{interval}
\usepackage[per-mode = symbol]{siunitx}
\usepackage{booktabs, multirow}
\usepackage{caption, subcaption}

\usepackage{algorithm}
\usepackage{algpseudocode}
\usepackage{dcolumn}
\newcolumntype{d}[1]{D{.}{.}{#1} }

\usepackage{pgf}
\usepackage{tikz}
\usetikzlibrary{arrows.meta}

\usetikzlibrary{arrows}
\usetikzlibrary{calc}
\usetikzlibrary{shapes}
\usetikzlibrary{trees}
\usetikzlibrary{patterns}
\tikzset{>=stealth'}

\usetikzlibrary{automata}

\makeatletter
\renewcommand{\todo}[2][]{%
  \@todo[fancyline,caption={#2}, #1]{
    \begin{minipage}{\linewidth}
      \footnotesize #2
    \end{minipage}
  }%
}
\makeatother
\newcommand{\lcs}[1]{LCS#1}
\newcommand{\suprb}[1]{SupRB#1}

\begin{document}
\title{Investigating the Impact of Independent Rule Fitnesses in a Learning Classifier System}
\titlerunning{Independent Rule Fitnesses in an LCS}
% If the paper title is too long for the running head, you can set
% an abbreviated paper title here
%
\author{Michael Heider \inst{1} \orcidID{0000-0003-3140-1993} \and
Helena Stegherr \inst{1} \orcidID{0000-0001-7871-7309} \and
Jonathan Wurth \inst{1} \orcidID{0000-0002-5799-024X} \and
Roman Sraj \inst{1} \and
J\"org H\"ahner \inst{1} \orcidID{0000-0003-0107-264X}}
\authorrunning{Heider et al.}
% First names are abbreviated in the running head.
% If there are more than two authors, 'et al.' is used.
%
\institute{Universi\"at Augsburg, Am Technologiezentrum 8, Augsburg, Germany
\email{\{first\_name.last\_name\}@uni-a.de}}
\maketitle              % typeset the header of the contribution
%
%%
%% end of the preamble, start of the body of the document source.

\begin{abstract}
Achieving at least some level of explainability requires complex analyses for many machine learning systems, such as common black-box models. %while other systems profit from being, to a reasonable extent, inherently interpretable.
We recently proposed a new rule-based learning system, \suprb{}, to construct compact, interpretable and transparent models by utilizing separate optimizers for the model selection tasks concerning rule discovery and rule set composition.
This allows users to specifically tailor their model structure to fulfil use-case specific explainability requirements. 
From an optimization perspective, this allows us to define clearer goals and we find that---in contrast to many state of the art systems---this allows us to keep rule fitnesses independent.
In this paper we investigate this system's performance thoroughly on a set of regression problems and compare it against XCSF, a prominent rule-based learning system.
We find the overall results of \suprb{'s} evaluation comparable to XCSF's while allowing easier control of model structure and showing a substantially smaller sensitivity to random seeds and data splits.
This increased control can aid in subsequently providing explanations for both training and final structure of the model.
\end{abstract}

%%
%% Keywords. The author(s) should pick words that accurately describe
%% the work being presented. Separate the keywords with commas.

\section{Introduction}
The applicability of decision making agents utilizing machine learning methods in real-world scenarios depends not only on the accuracy of the models, but equally on the degree to which explanations of the decisions can be provided to the human stakeholders.
For example, in an industrial setting, experienced machine operators often rather rely on their own knowledge instead of on---in their eyes---unsubstantiated recommendations of the model going against that knowledge.
This problem is exacerbated as it is inevitable that the model is not perfect in every detail, especially when the learning task is complex and the available training data limited.

To still make use of the advantages of recommendations made by digital agents, increasing the trust of stakeholders in the predictions is essential.
It includes providing explanations of the processes involved to produce these, as well as of the entire model.
This can get to a point where easily explainable models are preferred over better performance with higher complexity. 
Rule-based learners such as Learning Classifier Systems (\lcs{s}) are well suited in these settings as they facilitate extensive explanations. \cite{Heider21}

\lcs{s} \cite{urbanowicz2009} are inherently transparent and interpretable rule-based learners that make use of a finite set of if-then rules to compose their models.
Each rule contains a simpler, more comprehensible submodel, related to specific areas of the feature space.
The conditions under which rules apply are optimized during the training process, commonly by an evolutionary algorithm.
There are two main styles of \lcs{s}: Pittsburgh-style systems, which evolve a population of sets of rules with combined fitnesses (one per set), and Michigan-style systems, which adapt a single set of rules over time with individual fitnesses (one per rule).
Therefore, optimization by the evolutionary algorithm is performed differently in the two styles, but always aimed at finding an ``accurate and maximally general'' \cite{urbanowicz2017} set of rules.
Explainability requisites are commonly not directly included as optimization targets for the much more frequent Michigan-style systems, though it is to some extent represented under the concept of generality.
In Pittsburgh-style systems, the evolutionary algorithm does typically include error and rule set size as targets but it has to optimize the positioning and also the selection of rules.
Therefore, each iteration is comprised of several changes to rules in the set which leads to common situations where beneficial changes to a rule are not reflected in a corresponding change to the fitness of the set and might therefore be discarded for the next generation.
While the suboptimal positioning of rules might not even decrease the system's performance, it is, however, a problem when explanations concerning the rule conditions or the training process should be given.
Michigan-style systems, on the other hand, often generate and keep a large set of both good and suboptimal rules, in total, far more than required for the given problem.
Therefore, they need additional procedures after training, especially \emph{compaction} techniques, to reduce the population to the most important rules and therefore to enhance explainability \cite{tan2013,liu2021b}.

The first description of a new \lcs{} algorithm, in which the optimization of rule conditions is separated from the composition of rules to form a problem solution, was provided in \cite{heider2022}.
This way, rule fitnesses are kept independent from other influences than their direct changes, increasing the \emph{locality}.
It also improves the explainability of these quality parameters.
Additionally, explainability is improved through the direct control over population sizes and whether good rules should be optimized to be more specific or more general.
In this paper, we extend the initial examinations of \suprb{}, as described in \Cref{suprb2}, by evaluating against a modern version of XCSF \cite{wilson2002,Preen2021}, one of the most developed and advanced \lcs{s}, on a variety of different regression datasets (cf. \Cref{eval}).
We find that, as intended, \suprb{} performs competitively based on hypothesis testing on error distributions as well as Bayesian comparison \cite{benavoli2017} across datasets, while producing more compact models directly.

\section{Related Work}
\label{related}

XCS is a prominent representative of \lcs{s}.
Its many derivatives and extensions are capable of solving all three major learning tasks \cite{urbanowicz2009}.
In the context of this paper, the most notable extensions are those concerned with applicability to real-valued problem domains and supervised function approximation.
In terms of real-valued problem domains, this means replacing binary matching function with interval-based ones \cite{wilson2000a}.
For supervised function approximation, XCSF was designed \cite{wilson2002}.
It replaces the constant predicted payoff with a linear function.
To further enhance the performance, more complex variants were introduced to replace linear models and interval-based matching functions \cite{bull2002b,lanzi2006}, however, at the cost of overall model transparency.

\lcs{s} are commonly considered as transparent or interpretable by design, as are other rule-based learning systems, and naturally relate to human behaviour.
In contrast, other systems require extensive post-hoc methods, such as visualisation or model transformation, to reach explainability.
Even though \lcs{} can be seen as inherently transparent, there can be factors that reduce these capabilities.
They may arise through the encodings used, the number of rules in general and the complexity introduced by using complex matching functions or submodels in the individual rules. \cite{barredoArrieta2020}

Controlling these limitations in \lcs{s} is typically done by design but can incorporate designated post-hoc methods.
Post-hoc methods, especially visualisation techniques for classifiers, can improve the interpretability of the model \cite{urbanowicz2012,liu2019,liu2021}.
However, they have to be devised or adapted to the specific needs of the problem at hand and the model itself, which requires time and expertise. 
Controlling transparency by design can therefore be beneficial in some cases.
While some factors, for example problem-dependent complex variables/features, restrict interpretability and can hardly be influenced, other factors can compensate for these issues.
This means the design must consider understandable matching functions and predictive submodels, without foregoing an adequate predictive power.

Another aspect strongly related to the interpretability of \lcs{} models is the size of the resulting rule sets, e.g.\@ smaller sets facilitate direct visual inspection and require less subsequent analysis.
Controlling this size is handled differently in Pittsburgh-style and Michigan-style systems.
Pittsburgh-style \lcs{s} utilize the fitness function of the optimization algorithm, which often incorporates different objectives, i.e.\@ accuracy and number of rules.
A prominent example is GAssist \cite{bacardit2004}, where accuracy and minimum description length form a combined objective and an additional penalty is given if the rule set size gets too small.
Michigan-style systems, on the other hand, do not control the rule set size by means of the fitness function, as large populations are often beneficial for the training process.
During the training, subsumption can be performed to merge two rules where one fully encompasses the other.
Compaction is a post-hoc method to reduce the size of the rule set after training by removing redundant rules without decreasing the prediction accuracy \cite{wilson2002a,liu2021b}.
However, most compaction methods are purely designed for classification.

\section{The Supervised Rule-based Learning System}
\label{suprb2}

We recently proposed \cite{heider2022} a new type of LCS with interchanging phases of rule discovery and solution composition, the Supervised Rule-based Learning System (\suprb{}).
The first phase optimizes rule conditions independently of other rules, discovering a diverse pool of well proportioned rules.
Subsequently, in the second phase, another optimization process selects a subset of all available rules to compose a good (accurate yet small) solution to the learning task.
In contrast to other LCSs, we thus separate the \emph{model selection} objectives of finding multiple well positioned rules (with a tradeoff between local prediction error and matched volume) and selecting a set of these rules for our final model.
That allows us to predict arbitrary inputs with minimal error while the set of rules is as small as possible to keep transparency and interpretability high.
As it can be difficult to determine how many rules would need to be generated before a good solution can be composed from them, the two phases are alternated until some termination criterion, e.g.\@ a certain number of iterations, is reached (cf.\@ \Cref{alg:mainloop}).
Note that, in contrast to Pittsburgh-style systems, rules added to the pool remain unchanged and will not be removed throughout the training process.
An advantage of alternating phases is the ability to steer subsequent rule discoveries towards exploring regions where no or ill-placed rules are found, based on information from the solution composition phase.

\begin{algorithm}
\caption{\suprb{'s} main loop}\label{alg:mainloop}
\begin{algorithmic}[1]
\State $\text{pool}\gets \emptyset$
\State elitist $\gets \emptyset$
%\State $i\gets 0$
%\While{$i < \text{n\_iter}$}
\For{$i\gets 1, \text{n\_iter}$}
\State $\text{pool}\gets\text{pool}\cup\Call{discover rules}{\text{elitist}}$
\State $\text{elitist}\gets\Call{compose solution}{\text{pool}, \text{elitist}}$
%\State $i\gets i+1$
%\EndWhile
\EndFor
\State \textbf{return} elitist
\end{algorithmic}
\end{algorithm}

Insights into decisions are a central aspect of \suprb{}, therefore, its model is kept as simple and interpretable as possible \cite{heider2022}:
\begin{enumerate}
\item Rules' conditions use an interval based matching: 
A rule $k$ applies for example $x$ iff $x_i \in [l_{k, i}, u_{k, i}] \,\forall i$ with $l$ being the lower and $u$ the upper bounds.
\item Rules' submodels $f_k(x)$ are linear.
They are fit using linear least squares with a l2-norm regularization (Ridge Regression) on the subsample matched by the respective rule.
\item When mixing multiple rules to make a prediction, a rule's experience (the number of examples matched during training and therefore included in fitting the submodel) and in-sample error are used in a weighted sum.
\end{enumerate}

\begin{algorithm}
\caption{\suprb{'s} Rule Discovery}\label{alg:RD}
\begin{algorithmic}[1]
\Procedure{discover rules}{elitist}
\State rules $\gets \emptyset$
\For{$i\gets 1, \text{n\_rules}$} \Comment{($1,\lambda$)-ES for each new rule}
%\State $j\gets 0$
%\State $\text{best\_children} \gets [\,]$
\State candidate, proponent $\gets$ \Call{init rule}{elitist} 
\Repeat
\State $\text{children}\gets \emptyset$
\For{$k\gets 1,\lambda$}
\State children $\gets \text{children}\, \cup\,$\Call{mutate}{proponent}
\EndFor
\State proponent $\gets$ child with highest fitness
%\State Add proponent to best\_children
\If{candidate's fitness $<$ proponent's fitness}
\State candidate $\gets$ proponent
\State $j\gets 0$
\Else
\State $j\gets j+1$
\EndIf
\Until{$j=\delta$}
%\Until{$\text{best\_children}[j-\delta]_{fitness} > \text{best\_children}[k]_{fitness}; \forall k \in \mathbb{N}, k > j-delta, k \leq j $}
\State $\text{rules}\gets \text{rules}\cup\text{candidate}$
\EndFor
\State \textbf{return} {rules}
\EndProcedure
\end{algorithmic}
\end{algorithm}

In general, a large variety of methods can be used to discover new rules, but for this paper, we utilize an evolution strategy (ES).
The overall process is displayed in \Cref{alg:RD}.
While during a rule discovery phase typically multiple rules are discovered and added, this happens independently (and can be parallelized) in multiple $(1,\lambda)$-ES runs.
The initial candidate and parent rule is placed around a roulette-wheel selected training example, assigning higher probabilities to examples whose prediction showed a high in-sample error in the current (intermediate) solution (or \emph{elitist}).
The non-adaptive mutation operator samples a halfnormal distribution twice per dimension to move the parent's upper and lower bounds further from the center by the respective values.
This is repeated to create $\lambda$ children.
From these, the fittest individual is selected based on its in-sample error and the matched feature space volume as the new parent.
If it displays a higher fitness than the candidate it becomes the new candidate.
Specifically, the fitness is calculated as 
\begin{equation}
  \label{eq:fitness}
  F(o_1, o_2) = \frac{(1 + \alpha^2) \cdot o_1 \cdot o_2}{\alpha^2 \cdot o_1 + o_2} \,,
\end{equation}
with
\begin{equation}
  \label{eq:pseudo-accuracy}
  o_1 = \text{PACC} = \exp(-\text{MSE} \cdot \beta) \,,
\end{equation}
and
\begin{equation}
\label{eq:volume-share}
  o_2 = V = \prod_{i}{\frac{u_{i} - l_{i}}{\min_{x \in \mathcal{X}}{x_i} - \max_{x \in \mathcal{X}}{x_i}}} \,.
\end{equation} 
The base form (cf.\@ \cref{eq:fitness}) was adapted from~\cite{wu2019}, where it was combining two objectives in a feature selection context. 
The Pseudo-Accuracy (PACC), \cref{eq:pseudo-accuracy}, squashes the Mean Squared Error (MSE) of a rule's prediction into a $(0,\,1]$ range, while the volume share $V \in \interval{0}{1}$ (cf.\@ \cref{eq:volume-share}) of its bounds is used as a generality measure.
The parameter $\beta$ controls the slope of the PACC and $\alpha$ weighs the importance of $o_1$ against $o_2$.
We tested multiple values for $\beta$ and found $\beta = 2$ to be a suitable default.
For $\alpha$, 0.05 can be used in many problems (hyperparameter tuning for the datasets in this paper selected it in 3 out of 4 cases) but, ultimately, the value should always depend on the model size requirements, which are task dependent.
%Maximizing both objectives hence corresponds to generating rules that have minimal error and are maximally general.
%A special form of plus-selection is used in the ES, which simultaneously controls the number of iterations: for every iteration, the best of $\lambda$ children is saved as an elitist and compared with all elitists from previous iterations. 
If the candidate has not changed for $\delta$ generations, the optimization process is stopped and this specific elitist is added to the pool.
This process of discovering a new rule and adding it to the pool of rules is repeated until the set number of rules has been found.
We want to stress that this optimizer is not meant to find a single globally optimal rule as in typical optimization problems, but rather find optimally placed rules so that for all inputs a prediction can be made that is more accurate than a trivial model, i.e.\@ simply returning the mean of all data.
Therefore, independent evolution is advantageous.

\begin{algorithm}
\caption{\suprb{'s} Solution Composition}\label{alg:SC}
\begin{algorithmic}[1]
\Procedure{compose solution}{pool, elitist}
\State $\text{population}\gets \text{elitist}$
\For{$i\gets 1,\text{pop\_size}$}
\State $\text{population}\gets \text{population}\cup\Call{init solution}$
\EndFor
\For{$i\gets 1, \text{generations}$}
\State $\text{elitists}\gets \Call{select elitists}{\text{population}}$
\State parents $\gets \Call{tournament selection}{\text{population}}$
\State children $\gets \Call{crossover}{\text{parents}} $\Comment{90\% probability n-point}
\State population $\gets \Call{mutate}{\text{children}}$ \Comment{probabilistic bitflip}
\State $\text{population}\gets \text{population}\cup \text{elitists}$
\EndFor
\State \textbf{return} best solution from population
\EndProcedure
\end{algorithmic}
\end{algorithm}

In the solution composition phase, a genetic algorithm (GA) selects a subset of rules from the pool to form a new solution.
As with the rule discovery, many optimizers could be used and a few have already been tested in \cite{wurth2022}, finding that the GA is a suitable choice.
Solutions are represented as bit strings, signalling whether a rule from the pool is part of the solution.
The GA uses tournament selection to select groups of two solutions and combines two parents by using $n$-point crossover with a default crossover probability of 90\%.
Then, mutation is applied to the children, flipping each bit with a probability determined by the mutation rate.
The children and some of the fittest parents (\emph{elitism}) form the new population. 
The number of elitists depends on the population size of the GA, but in our experiments, we found 5 or 6 to work best with a population size of 32.
Solution fitness is also based on \cref{eq:fitness}.
Here, the solution's in-sample mean squared error and its \emph{complexity}, i.e.\@ the number of rules selected, are used as first and second objective, respectively.
Note that each individual in the GA always corresponds to a subset of the pool. 
Rules that are not part of the pool can not be part of a solution candidate and rules remain unchanged by the GA's operations.

\suprb{} is conceptualised and designed as a regressor.
This is reflected in both the description above and the evaluation in the following section.
However, we want to propose how the system could be adapted easily towards solving classification problems:
The linear submodels would need to be replaced with an appropriate classifier, either simply a constant model, logistic regression or a more complex model if the explainability requirements allowed that.
Additionally, the fitness functions would need to use accuracy (or an appropriate scoring for imbalanced data) instead of PACC and MSE.

\section{Evaluation}
\label{eval}

For our evaluation of the proposed system, we compare \suprb{} to a recent XCSF\footnote{\url{https://github.com/rpreen/xcsf} $\qquad\;$ \url{https://doi.org/10.5281/zenodo.5806708}} \cite{wilson2002,Preen2021}
with hyperrectangular conditions and linear submodels (with recursive least squares updates \cite{lanzi2006b}), as they closely correspond to the conditions and submodels used in \suprb{}. 
We acknowledge that some better performing conditions, e.g.\@ hyperellipsoids \cite{Butz2005a}, have been proposed for XCSF, however, we consider them less interpretable in high dimensional space for the average user.

\subsection{Experiment Design}
\label{design}

\suprb{} is implemented\footnote{\url{https://github.com/heidmic/suprb} $\quad$ \url{https://doi.org/10.5281/zenodo.6460701}} in Python 3.9, adhering to \emph{scikit-learn} \cite{pedregosa2011scikit-learn} conventions. 
Input features are transformed into the range $\interval{-1}{1}$, while the target is standardized.
Both transformations are reversible but improve \suprb{'s} training process as they help preventing rules to be placed in regions where no sample could be matched and remove the need to tune error coefficients in fitness calculations, respectively.
Based on our assumptions about the number of rules needed, 32 cycles of alternating rule discovery and solution composition are performed, generating four rules in each cycle for a total of 128 rules.
For the ES we selected a $\lambda$ of 20.
Additionally, the GA is configured to perform 32 iterations with a population size of 32.
To tune some of the more sensitive parameters, we performed a hyperparameter search using a Tree-structured Parzen Estimator in the Optuna framework~\cite{akiba2019} that optimizes average solution fitness on 4-fold cross validation.
We tuned datasets independently for 256 iterations per tuning process.
For XCSF we followed the same process, selecting typical default values\footnote{\url{https://github.com/rpreen/xcsf/wiki/Python-Library-Usage}} \cite{Preen2021} and tuning the remaining parameters independently on the four datasets using the same setup as before. 
The final evaluation, for which we report results in \Cref{results}, uses 8-split Monte Carlo cross-validation, each with \SI{25}{\percent} of samples reserved as a validation set.
Each learning algorithm is evaluated with 8 different random seeds for each 8-split cross-validation, resulting in a total of 64 runs.

%\begin{table}[htp]
%  \centering
%  \caption{\textbf{Overview of the datasets \suprb{} and XCSF are compared on.}}
%  \label{tab:datasets}
%  \begin{tabular}{lrr}
%    \toprule
%    {Name (Abbreviation)}                          & {$n_{\text{dim}}$} & {$n_{\text{sample}}$} \\
%    \midrule
%    Combined Cycle Power Plant (CCPP) & 4         & \num{9568}       \\
%    Airfoil Self-Noise (ASN)             & 5            & \num{1503}       \\
%    Concrete Strength (CS)                & 8            & \num{1030}       \\
%    Energy Efficiency Cooling (EEC)              & 8            & \num{768}        \\
%    \bottomrule
%  \end{tabular}
%\end{table}

We evaluate on four datasets part of the UCI Machine Learning Repository~\cite{dua2017uci}.
%An overview of number of sample size and dimensionality is given in \Cref{tab:datasets}.
The Combined Cycle Power Plant (CCPP)~\cite{kaya2012ccpp,tufekci2014ccpp} dataset shows an almost linear relation between features and targets and can be acceptably accurately predicted using a single rule.
Airfoil Self-Noise (ASN)~\cite{brooks1989airfoil} and Concrete Strength (CS)~\cite{yeh1998concrete} are both highly non-linear and will likely need more rules to predict the target sufficiently.
The CS dataset has more input features than ASN but is easier to predict overall.
Energy Efficiency Cooling (EEC)~\cite{tsanas2012energy} is another rather linear dataset, but has a much higher input features to samples ratio compared to CCPP.
It should similarly be possible to model it using only few rules.

\subsection{Results}
\label{results}

In our experiments we find that XCSF and \suprb{} achieve comparable results.
\Cref{tab:results} presents the dataset-specific performance in detail.
All entries are calculated on 64 runs per dataset (cf. \Cref{design}). 
As both systems were trained for standardized targets, we denote the results for the mean (across runs) mean squared errors (MSE) and their standard deviation (STD) as $\text{MSE}_\sigma$ and $\text{STD}_\sigma$, respectively.
Standardized targets allow better comparison between the datasets as results are on a more similar scale. 
Additionally, as many real world datasets are normally distributed, this should lighten the need to carefully hand tune the balance between solution complexity and error.
Note that predictions of both models can always be retransformed into the original domain.
Subsequently, $\text{MSE}_\text{orig}$ references the mean MSE in units of the original dataset-specific target domain.
Although this column is less helpful for cross dataset performance interpretations, it allows comparison to other works on the same data.
We found that, on two datasets (CCPP and ASN), XCSF shows a better performance, albeit only slightly for CCPP, that can be confirmed through hypothesis testing (Wilcoxon signed-rank test using a confidence level of 5\%).
Contrastingly, for the CS dataset, the hypothesis could not be rejected.
Thus, although \suprb{} shows a slightly lower mean MSE, this is not statistically significant. 
For the EEC dataset \suprb{} outperforms XCSF.

\begin{figure*}[t]
  \centering
  \begin{subfigure}[b]{0.47\linewidth}
    \centering
    \includegraphics[width=\linewidth]{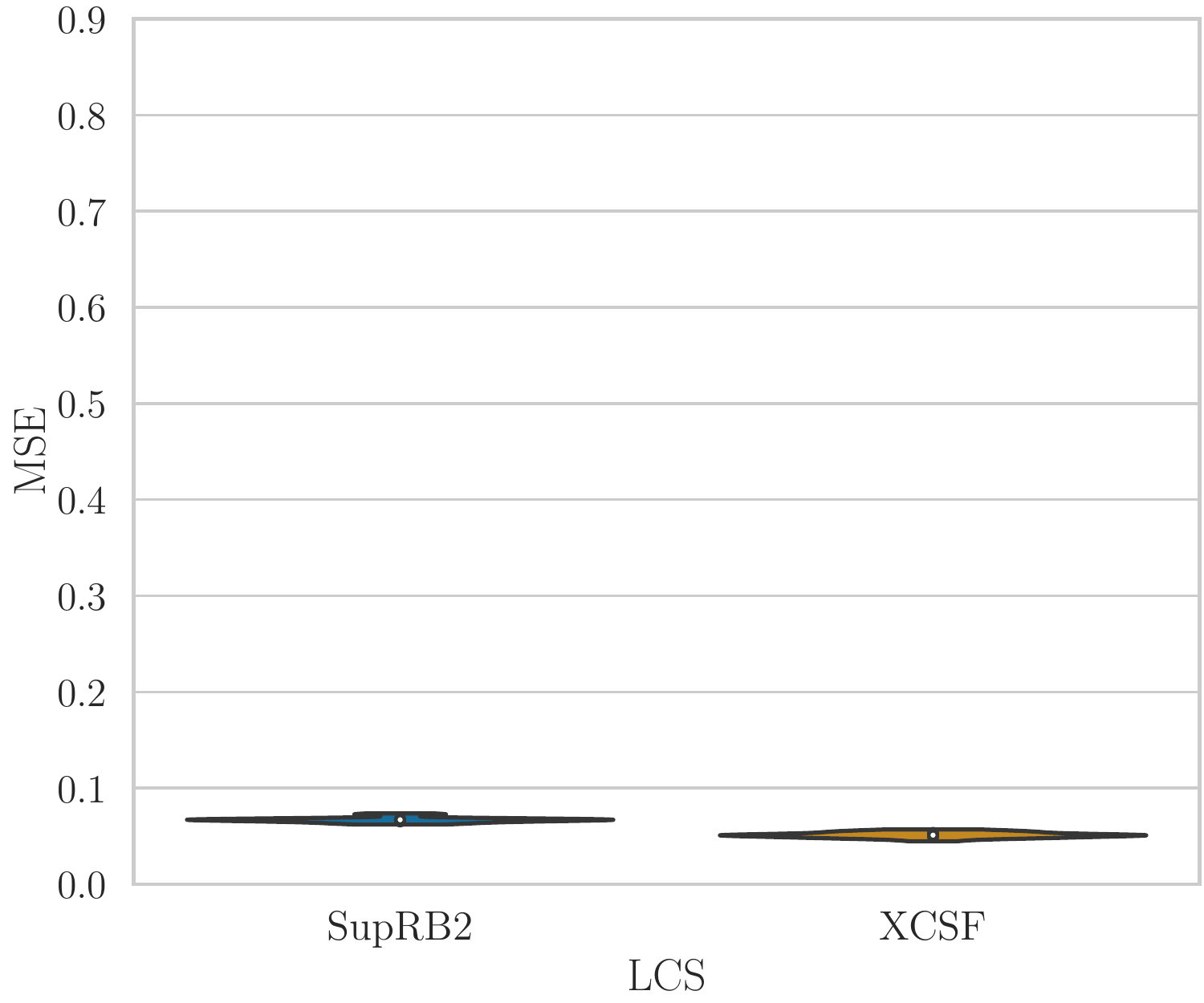}
    \subcaption{Distribution of runs on CCPP}
    \label{fig:ccpp_dist}
  \end{subfigure}
  \begin{subfigure}[b]{0.47\linewidth}
    \centering
    \includegraphics[width=\linewidth]{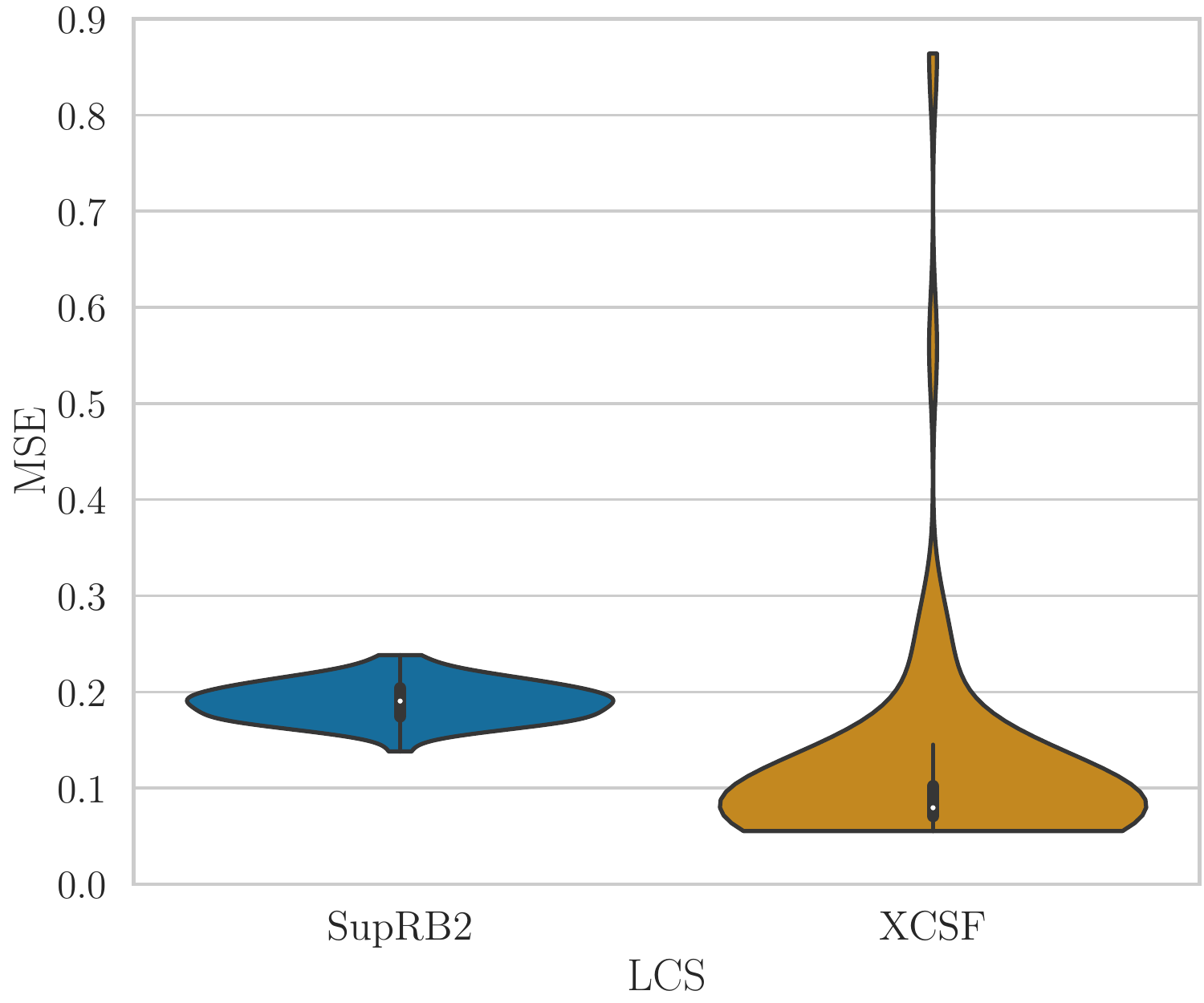}
    \subcaption{Distribution of runs on ASN}
    \label{fig:asn_dist}
  \end{subfigure}
  \begin{subfigure}[b]{0.47\linewidth}
    \centering
    \includegraphics[width=\linewidth]{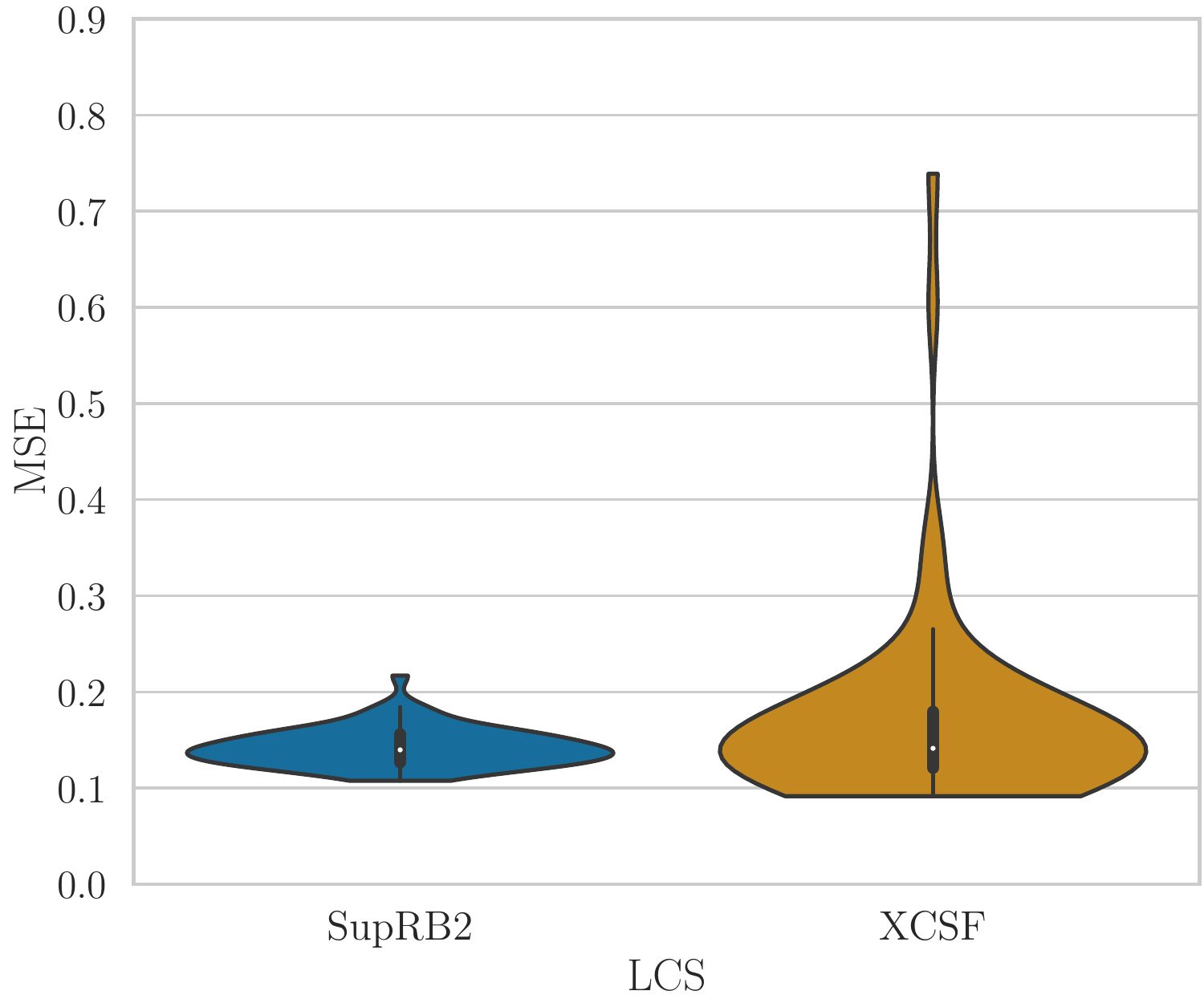}
    \subcaption{Distribution of runs on CS}
    \label{fig:cs_dist}
  \end{subfigure}
  \begin{subfigure}[b]{0.47\linewidth}
    \centering
    \includegraphics[width=\linewidth]{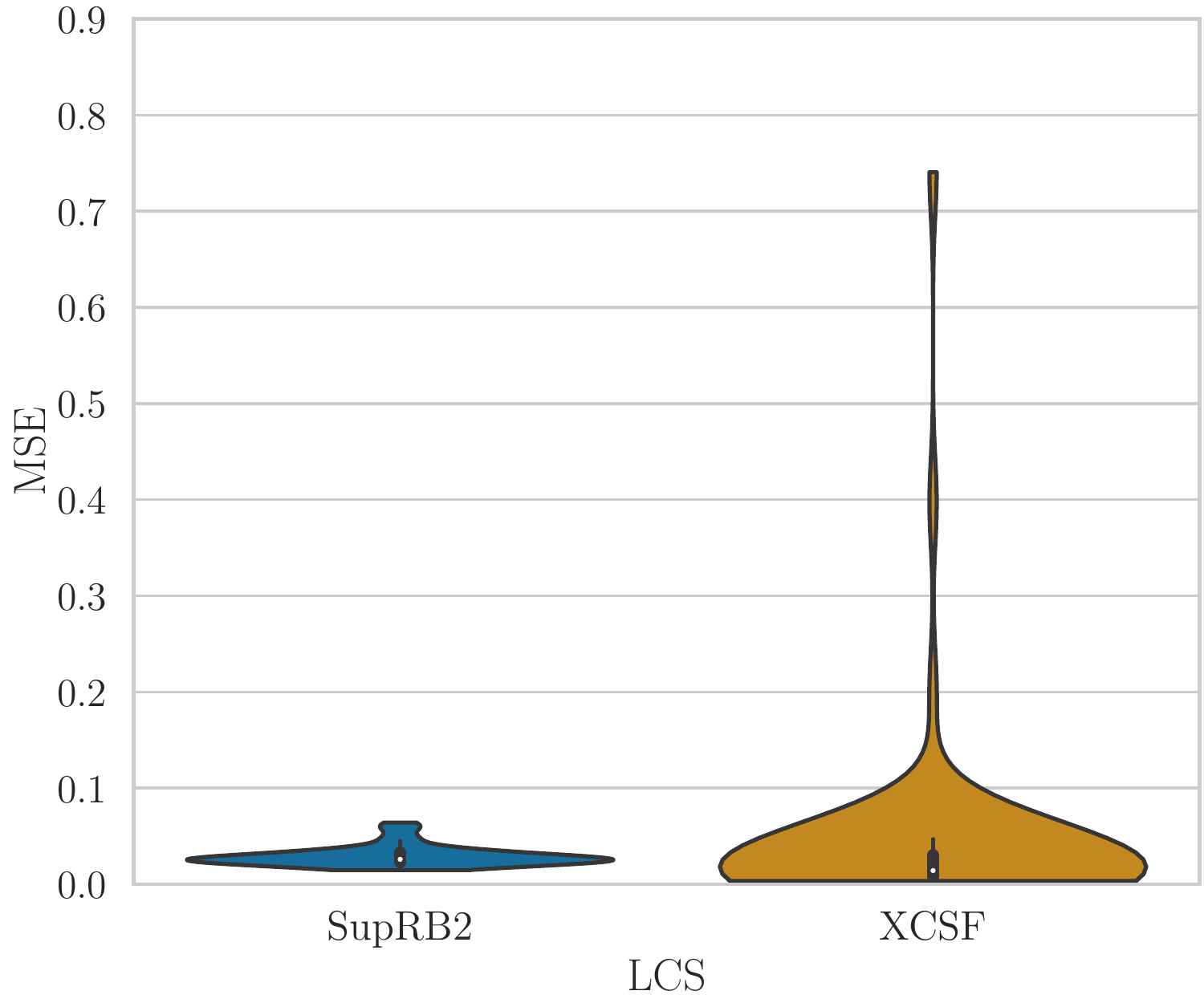}
    \subcaption{Distribution of runs on EEC}
    \label{fig:eec_dist}
  \end{subfigure}
  \caption{\textbf{Distribution of runs' errors on an equal scale.}}
  \label{fig:dists}
\end{figure*}

\begin{table}[t]
  \centering
  \caption{\textbf{Overview of the experimental test data results of 64 runs per dataset rounded to four decimals.}  \textmd{$\text{MSE}_\text{orig}$ and $\text{MSE}_\sigma$ give the means of the mean squared errors (MSE) in the dataset's original or a standardised target space, respectively. Similarly, $\text{STD}_\sigma$ displays the standard deviation of MSEs in standardised space. Highlighted in bold are the models where a 5\% significance Wilcoxon signed-rank test rejected the null hypothesis of equivalent distributions and the mean was better.}}
  \label{tab:results}
  \setlength{\tabcolsep}{5pt}
  \begin{tabular}{rrrrrrrr}
    \toprule
     & \multicolumn{3}{c}{CCPP} & $\:$ &\multicolumn{3}{c}{ASN} \\
   % \cmidrule(lr){2-13}
    \cmidrule{2-4}
    \cmidrule{6-8}
    & $\text{MSE}_\text{orig}$ & $\text{MSE}_\sigma$ & $\text{STD}_\sigma$ && $\text{MSE}_\text{orig}$ & $\text{MSE}_\sigma$ & $\text{STD}_\sigma$ \\
    \cmidrule{2-4}
    \cmidrule{6-8}
    %\midrule
    XCSF & \textbf{0.8745} & \textbf{0.0512} & \textbf{0.0028} && \textbf{0.7930} & \textbf{0.1150} & \textbf{0.1195} \\
    SupRB & 1.1433 & $0.0669$ & $0.0027$ && 1.3079 & $0.1896$ & $0.0199$ \\
    \cmidrule{2-4}
    \cmidrule{6-8}\\
    \noalign{\vskip -3mm}
    &\multicolumn{3}{c}{CS} &&\multicolumn{3}{c}{EEC} \\
    \cmidrule{2-4}
    \cmidrule{6-8}
    & $\text{MSE}_\text{orig}$ & $\text{MSE}_\sigma$ & $\text{STD}_\sigma$ && $\text{MSE}_\text{orig}$ & $\text{MSE}_\sigma$ & $\text{STD}_\sigma$ \\
%    \cmidrule(lr){2-4}
%    \cmidrule(lr){5-7}
    \cmidrule{2-4}
    \cmidrule{6-8}
    XCSF & 2.8291 & $0.1694$ & $0.1043$ && 0.3660 & $0.0385$ & $0.1032$ \\
    SupRB & 2.3779 & $0.1424$ & $0.0199$ && \textbf{0.2776} & \textbf{0.0292} & \textbf{0.0107} \\
    \bottomrule
  \end{tabular}
\end{table}

%\begin{figure}
%\centering
%    \includegraphics[width=.8\linewidth]{figs/combined_cycle_power_plant_error_zoomed}
%    \caption{Zoomed in version of \Cref{fig:ccpp_dist}}
%    \label{fig:ccpp_dist_zoomed}
%\end{figure}

%\michiin{eventuell will man das ein bisschen kürzen, vor allem, wenn die zweite figure rausfällt muss der Text etwas anders sein}
We found that \suprb{'s} runs had a similar (to each other) performance much more consistently than XCSF's.
This is shown by $\text{STD}_\sigma$ (cf. \Cref{tab:results}) and specifically illustrated in \Cref{fig:dists}, which shows the distribution of test errors across all 64 runs.
For three of the four datasets, XCSF shows some strong outliers that go against its remaining performances.
Additionally, the majority of runs is also further distributed around the mean and median values.
We assume that this is largely due to the stochastic iterative nature of training in XCSF. %\michi{cover delete cycle? kann man hier die shared fitness blamen?}
For the CCPP dataset (\Cref{fig:ccpp_dist}) no outliers were produced by XCSF and overall performance is quite similar across runs. 
This is especially noticeable when comparing the distribution to those on the other datasets. 
In fact, the runs are so similar (even across models) that it is hard to make any analysis on this scale.
%\Cref{fig:ccpp_dist_zoomed} shows the results on CCPP zoomed in as far as possible.
%Even here, the variance is clearly low ($\text{STD}_\text{XCSF}$: 0.0028; $\text{STD}_\text{\suprb{}}$: 0.0027), but XCSF outperformed \suprb{} on average. \michi{very slightly and (probably) practically insignificant}
Although, XCSF slightly outperformed \suprb{} on average on CCPP, as confirmed by statistical testing, we can assume that this advantage is likely not practically significant.
From a graphical perspective (cf. \Cref{fig:cs_dist}), \suprb{} seems to produce more desirable models on CS, even if the hypothesis testing remained ambiguous.
On EEC, XCSF achieves a slightly better median MSE performance ($\text{Median}_\text{XCSF}$: 0.014; $\text{Median}_\text{\suprb{}}$: 0.026), however, its mean MSE is poorer due to badly performing runs.
Regardless, the overall performance can be viewed as rather close, although both sets of runs are clearly not following the same distribution.
As \suprb{'s} and XCSF's models were trained on the same random seeds and cross-validation splits, we can conclude that \suprb{} is overall more reliable even if not necessarily better.
%\michi{we probably want to explain a bit more why we think this happens. possibly/likely due to how stupid nicheing is}

\begin{table}[ht]
  \centering
  \caption{\textbf{Overview of the solution complexities} (number of rules in the solution proposed by \suprb{} or the final macro-classifier count in an XCSF population, respectively) across 64 runs per dataset.}
  \label{tab:complexities}
  \setlength{\tabcolsep}{5pt}
%  \begin{tabular}{l d{2} d{2} d{2} d{2} r d{2} d{2} d{2} d{2}}
\begin{tabular}{l rrrrrrrrr}
    \toprule
    & \multicolumn{4}{c}{\suprb{}} && \multicolumn{4}{c}{XCSF} \\
    \cmidrule{2-5}
    \cmidrule{7-10}
     & \multicolumn{1}{l}{CCPP} & \multicolumn{1}{l}{ASN} & \multicolumn{1}{l}{CS} & \multicolumn{1}{l}{EEC} &&  \multicolumn{1}{l}{CCPP} & \multicolumn{1}{l}{ASN} & \multicolumn{1}{l}{CS} & \multicolumn{1}{l}{EEC}\\
     %\cmidrule(lr){3-6}
    %\cmidrule(lr){8-11}
    \cmidrule{2-5}
    \cmidrule{7-10}
%    mean && 2.65 & 26.42 & 22.31 & 12.81     && 2496.11 & 1045.12 & 623.13 & 1137.13\\
%    st. dev. && 0.62 & 2.47& 2.60 & 1.71    && 3.1 & 3.15 & 2.98 & 3.4\\
%    median && 3 & 27 & 22& 13    && 2495 & 1044 & 622 & 1136\\
%    min && 2 & 19 & 17 & 9   && 2494 & 1041 & 622 & 1135\\
%    max && 4 & 30 & 30 & 17   && 2510 & 1059 & 631 & 1153\\
    mean & 2.65 & 26.42 & 22.31 & 12.81     && 2253.28 & 962.03 & 562.81 & 1028.78\\
    st. dev. & 0.62 & 2.47& 2.60 & 1.71    && 24.70 & 9.17 & 11.73 & 14.90
    \vspace{2pt}\\
    median & 3 & 27 & 22& 13    && 2250 & 962 & 562 & 1026\\
    min & 2 & 19 & 17 & 9   && 2202 & 934 & 530 & 994\\
    max & 4 & 30 & 30 & 17   && 2301 & 980 & 593 & 1068\\
    \bottomrule
  \end{tabular}
\end{table}

For \suprb{} we directly control the size (number of rules; \emph{complexity}) of the global solution via the corresponding fitness function used in the GA. 
\Cref{tab:complexities} shows the complexities of the 64 runs per dataset.
Note that the highest theoretical complexity is 128, as we did only add 128 rules to the pool.
We find that, although theoretically a single rule is able to predict CCPP well, the optimizer prefers to use at least two but at most four rules, achieving slightly better errors than with a singular linear model.
As expected, the solutions to the two highly non-linear datasets (ASN and CS) do feature considerably more rules.
EEC again was solved with fewer rules, speaking to its more linear nature, although with more than CCPP, for which a linear solution exists.
Standard deviations of complexities increase as the mean increases and the median stays close to the mean.

XCSF seems to have fallen into a cover-delete-cycle where rules did not stay part of the population for long. 
Covering is a rule generation mechanism that creates a new rule whenever there were too few matching rules.
The deletion mechanism removes rules when the population is too full, as there exists a hyperparameter-imposed maximum population size.
In our tuning, we did tune both the number of training steps and the maximum population size (among the many other parameters of XCSF) and find that post-training populations are at or around the maximum population size.
XCSF's hyperparameter tuning opted for much larger populations than the typical rule of thumb of using ten times as many rules as would be expected (from domain knowledge or prior modelling experience) for a good problem solution \cite{urbanowicz2017}.
Additionally, upon deeper inspection, we found that the rules were typically introduced late in the training process, however, the system error did not change in a meaningful manner long before that point.
%Note that, while we did not perform compaction on XCSF, which would make a direct comparison unfair as XCSF relies on this post-hoc technique to produce low complexity rule sets, we did utilize subsumption in the EA.
Note that we did utilize subsumption in the EA.
This mechanism prevents the addition of a newly produced rule to the population when it is fully engulfed by a parent rule and instead increases the parents numerosity parameter.
A rule with numerosity $n$ counts as $n$ rules with numerosity $1$ towards the maximum population size limit.
Subsumption thus theoretically decreases the actual number of classifiers in our population.
However, in our experiments the cover-delete-cycle seems to have rendered this mechanism useless.

It is reasonably possible that \suprb{'s} performance would improve in some cases if the pressure to evolve smaller rule sets was lower.
However, as explainability suffers with large rule sets, we think that the presented solutions strike an acceptable balance.
%Afterall, XCSF's solutions were substantially larger even if we account for rules with an experience of $0$ which would have been removed by most compaction methods.
Afterall, XCSF's solutions were substantially larger even after applying a simple compaction technique of removing rules with an experience of $0$ from the final population.
This compaction method removed on average about 10\% of rules from the run's populations.
\Cref{tab:complexities} reports the complexity results after compaction.
However, we acknowledge that a variety of compaction techniques exists for classification problems \cite{liu2021b} that could in some cases potentially be adjusted for the use within regression tasks.
Likely, SupRB and XCSF find themselves at different points on the Pareto front between error and complexity.
However, in \suprb{} we do not need to rely on additional post-processing but can solve this optimization problem directly and, importantly, balance the tradeoff of prediction error and rule set complexity against user needs, whereas compaction mechanisms are typically designed to decrease complexity only in a way as to not increase the \lcs{'s} system error \cite{liu2021b}.

%\michiin{definitely keep this. maybe fix the claims that infuriated David and the reviewers agreed with. if we repeat the xcsf pop size experiments, we could also compare with regards to complexity. surely not fair, but maybe good for us. we could spin the pareto front argument. in theory this gets harmed a bit when we didnt do compaction, though}
% (0.635, 0.0845, 0.2805)
Beyond dataset-specific performances, we would like to find a more general answer to the question whether the newly proposed \suprb{} does perform similarly to the well established XCSF.
This would indicate that we can find a good LCS model even without the niching mechanisms employed by XCSF's rule fitness assignment.
To find an initial answer based on the performed experiments we use a Bayesian model comparison approach \cite{benavoli2017} using a hierarchical model \cite{corani2017} that jointly analyses the cross-validation results across multiple random seeds and all four datasets.
We assume a region of practical equivalence of $0.01 \cdot \sigma_\text{dataset}$.
  \begin{align*}
p(\text{\suprb{}} \ll  \text{XCSF}) & \approx 63.4\,\%\\
p(\text{\suprb{}} \equiv  \text{XCSF}) & \approx \; 8.5\,\%\\
p(\text{\suprb{}} \gg  \text{XCSF}) & \approx 28.1\,\%
  \end{align*}
where:
\begin{itemize}
\item $p(\text{\suprb{}} \ll  \text{XCSF})$ denotes the probability that \suprb{} performs worse (achieving a higher MSE on test data),
\item $p(\text{\suprb{}} \equiv  \text{XCSF})$ denotes the probability that both systems achieve practically equivalent results and
\item $p(\text{\suprb{}} \gg  \text{XCSF})$ denotes the probability that \suprb{} performs better (achieving a lower MSE on test data).
\end{itemize} 

From these results we clearly can not make definitive assessments that XCSF is stronger than \suprb{}.
While it might outperform \suprb{} in less than two thirds of cases, it also will be outperformed in almost a third of cases.
\cite{benavoli2017} suggest thresholds of 0.95, 0.9 or 0.8 for probabilities to make automated decisions. 
The specific value needs to be chosen according to the given context.
We did not perform the same analysis for the rule set sizes as the results are quite clear with \suprb{} being the system very likely producing much smaller rule sets.
Overall, we can conclude that no clear decision can be made and that the newly developed (and to be improved in the future) \suprb{} should be considered an equal to the well established XCSF.

%\helenain{maybe now add rule example}
%\michiin{or we have a future work again?}
%\michiin{Cut---Example rule is much less important in this paper and is way too spacious anyway}

\begin{table*}[h]
\centering
\caption{Exemplary rule generated by \suprb{} on CS dataset. \textmd{The target is the concrete compressive strength. 
The original space intervals denote the area matched by the rule in terms of the original variable scales, while the intervals in feature spaces are scaled into $\interval{-1}{1}$ and help perceiving rule generality at a glance.
Coefficients denote the weight vector used for the linear model.}}
  \setlength{\tabcolsep}{5pt}
  \begin{tabular}{l r l r l r}
\toprule
  &$\quad$& \multicolumn{1}{c}{Original Space} &$\quad$& \multicolumn{2}{c}{Feature Space $\sigma$} \\
   \cmidrule(lr){3-3}
 \cmidrule(lr){5-6}
input variable && interval && interval & coefficient\\
 \cmidrule(lr){1-1}
   \cmidrule(lr){3-3}
 \cmidrule(lr){5-6}
    Cement  [$\si{\kg\per\cubic\m}$]            && $\interval{104.72}  {516.78}$ && $\interval{-0.99}{0.89}$ & 2.38 \\
    Blast Furnace Slag [$\si{\kg\per\cubic\m}$] && $\interval{0.00}       {359.40}$   & & $\interval{-1.00}{1.00}$ & 2.29 \\
    Fly Ash    [$\si{\kg\per\cubic\m}$]         && $\interval{13.45}   {200.0}$    && $\interval{-0.87}{1.00}$ & 0.68 \\
    Water      [$\si{\kg\per\cubic\m}$]         && $\interval{122.64}  {244.80}$ && $\interval{-0.99}{0.96}$ & -1.26 \\
    Superplasticizer  [$\si{\kg\per\cubic\m}$]  && $\interval{6.02}    {24.80}$  && $\interval{-0.63}{0.54}$ & -0.67 \\
    Coarse Aggregate [$\si{\kg\per\cubic\m}$]   && $\interval{950.16}  {1145.00}$   && $\interval{-0.13}{1.00}$ & 0.71 \\
    Fine Aggregate  [$\si{\kg\per\cubic\m}$]    && $\interval{756.14}  {992.60}$    && $\interval{-0.19}{1.00}$ & 0.60 \\
    Age      [days]           && $\interval{18.36}   {365.00}$  && $\interval{-0.90}{1.00}$ & 2.07 \\
 %\cmidrule(lr){2-2}
 \cmidrule(lr){5-6}
	&&&&\multicolumn{2}{c}{$\text{intercept}_\sigma = 3.9160$} \\
	\midrule
    \multicolumn{6}{l}{
    \begin{tabular}{l r ll r ll r}
     \multicolumn{1}{l}{In-sample $\text{MSE}_\text{orig}$} & \multicolumn{1}{r}{1.5310} & &
	 \multicolumn{1}{l}{In-sample $\text{MSE}_\sigma$} & \multicolumn{1}{r}{0.0917} & &
	 \multicolumn{1}{l}{Experience} & \multicolumn{1}{r}{84} \\
	\end{tabular}} \\
	\bottomrule
\end{tabular}
  \label{tab:example_rule}
\end{table*}

\Cref{tab:example_rule} presents a rule trained for the CS dataset.
It has an experience (number of matched examples during training) of 84 and matched another 31 examples during testing.
It is part of a model consisting of 23 rules with experiences of 7 to 240 with a mean experience of $54.17 \pm 55.63$.
The rules were, thus, either rather general or rather specific with this rule being on the more general side.
Upon closer inspection, for 5 of the 8 dimensions of CS the rule matches most of the available inputs (being maximally general on the ``Blast Furnace Slag'' input variable).
For the transformed input space (feature space) that is scaled to an interval of $\interval{-1}{1}$ this can easily be seen without any knowledge about the datasets structure, although it is likely that users of the model will have enough domain knowledge to be able to derive this directly from the intervals in the original space.
It can also be assumed that these users will generally prefer to inspect the rule in that representation.
%For the ``Superplasticizer'' input variable the rule matches slightly more than half of the possible values and is roughly positioned around the center of the range of values, while for ``Coarse Aggregate'' and ``Fine Aggregate'' the rule matches the 55 to 60 \% of the input range, oriented toward to higher values.
High concentrations of ``Water'' and ``Superplasticizer'' have negative effects on the compressive strength of the concrete for the aforementioned value ranges, while higher concentrations of ``Cement'', ``Blash Furnace Slag'' and ``Age'' of the mixture positively influence its compressive strength.
The other three input variables have positive but less pronounced effects.
%One of the examples matched by the rule is $x=(190.3,    0 ,  125.2,  161.9,    9.9, 1088.1,  802.6,   56)^T$ with $y=38.56$.
%For this example the rule proposes $\hat{y} = 38.39$.
%The prediction is clearly close to the ground truth.
%Providing this example to a user with domain experience can help them assess the model's predictive power.
%Moreover, the user can then extrapolate to other examples that would fit in the provided matching intervals, increasing their trust over other models, where such extrapolation is not possible from the models' structure.
%The rule performed below average compared to the other rules in this solution (mean in-sample MSE of 0.0751) on its training data, although this should be viewed critically as more specific rules tend to be able to fit their data more easily.
Overall, rule inspection offers some critical insights into the decision making process and can be done fairly easily based on the rule design and the low number of rules per solution.

\section{Conclusion}

In this paper, we expanded the view on the Supervised Rule-based Learning System (\suprb{}) with an optimization perspective.
We highlighted the advantages of individual rule fitnesses compared to the fitness-sharing approaches typical for other Learning Classifier Systems (LCSs) and discussed our approach to perform LCS model selection using two separated optimizers from that perspective.

To evaluate the system we compared it to XCSF, a well known LCS with a long research history, on four real world regression datasets with different dimensionalities and problem complexities.
As one of the greatest advantages of LCS compared to other learning systems is their inherent interpretability and transparency, we limited our study to the use of hyperrectangular conditions and linear models for both systems.
After hyperparameter searches for the more sensitive parameters (256 evaluations with 4-fold cross validation), we performed a total of 64 (8 random seeds and 8-fold cross validation with 25\% test data) runs of each system on every dataset.
We found that, in general, performance is relatively similar.
While XCSF showed a statistically better mean test error on two datasets, it was outperformed on one and no statistically significant decision could be made on the fourth dataset.
We performed a Bayesian model comparison approach using a hierarchical model and found that no clearly better model can be determined on errors.
Solution sizes of \suprb{} were better than XCSF's even when applying some form of compaction.
Additionally, \suprb{} was more consistent in its performance across runs.
Thus, we conclude that, for now and with future research pending, both systems produce similarly performing models.

\bibliographystyle{splncs04}
\bibliography{References}

\end{document}